\documentclass[sigconf]{acmart}

\copyrightyear{2021}
\acmYear{2021}
\setcopyright{iw3c2w3}
\acmConference[WWW '21 Companion]{Companion Proceedings of the Web Conference 2021}{April 19--23, 2021}{Ljubljana, Slovenia}
\acmBooktitle{Companion Proceedings of the Web Conference 2021 (WWW '21 Companion), April 19--23, 2021, Ljubljana, Slovenia}
\acmPrice{}
\acmDOI{10.1145/3442442.3451383}
\acmISBN{978-1-4503-8313-4/21/04}

\usepackage{float}
\usepackage{xcolor}

\usepackage{soul}
\begin{document}

\title{JSI at the FinSim-2 task: Ontology-Augmented\\ Financial Concept Classification }

\author{Timen Stepi\v{s}nik Perdih}
\affiliation{%
  \institution{Jo\v{z}ef Stefan Institute}
  \city{Ljubljana}
  \country{Slovenia}
}
\email{tstepisnikp@gmail.com}

\author{Senja Pollak}
\affiliation{%
  \institution{Jo\v{z}ef Stefan Institute}
  \city{Ljubljana}
  \country{Slovenia}}
\email{Senja.Pollak@ijs.si}

\author{Bla\v{z} \v{S}krlj}
\affiliation{%
  \institution{Jo\v{z}ef Stefan Institute and  Jo\v{z}ef Stefan International Postgraduate School}
  \city{Ljubljana}
  \country{Slovenia}
}
\email{blaz.skrlj@ijs.si}

\renewcommand{\shortauthors}{Timen Stepišnik Perdih, et al.}

\begin{abstract}
Ontologies are increasingly used for machine reasoning over the last few years. They can provide explanations of concepts or be used for concept classification if there exists a mapping from the desired labels to the relevant ontology. Another advantage of using ontologies is that they do not need a learning process, meaning that we do not need the train data or time before using them. This paper presents a practical use of an ontology for a classification problem from the financial domain. It first transforms a given ontology to a graph and proceeds with generalization with the aim to find common semantic descriptions of the input sets of financial concepts.


 We present a solution to the shared task on Learning Semantic Similarities for the Financial Domain (FinSim-2 task). The task is to design a system that can automatically classify concepts from the Financial domain into the most relevant hypernym concept in an external ontology - the Financial Industry Business Ontology. We propose a method that maps given concepts to the mentioned ontology and performs a graph search for the most relevant hypernyms. We also employ a word vectorization method and a machine learning classifier to supplement the method with a ranked list of labels for each concept. 
 \hl{The final version of this paper was published in the Proc. of The Web Conference: companion of the World Wide Web conference (WWW 2021): 30th edition, p. 298-301, doi: 10.1145/3442442.3451383.}
\end{abstract}


\maketitle

\section{Introduction}
With the increasing availability of domain ontologies, there is also a growing number of methods exploring and exploiting their use for different problems. Most frequently ontologies provide a source of domain knowledge usable by a computer, which inspires the creation of new systems that try to solve problems that only domain experts can address.

This work presents a method that uses an ontology to provide a solution to the shared task on Learning Semantic Similarities for the Financial Domain (FinSim-2 at FinWeb-2021), where the task is to classify concepts from the Financial domain into the most relevant hypernym concept in the \textbf{Financial Industry Business Ontology}\footnote{https://spec.edmcouncil.org/fibo/}. The solution consists of transforming the ontology into a NetworkX directed and unweighted graph \cite{NetworkX} and mapping concepts to the ontology. It then employs a directed graph search (generalization) to generalize and classify mapped concepts. The method's output is enriched by using Word2vec \cite{church_2017} to vectorize concepts and Random Forest Classifier~\cite{RF} to provide a ranked list of labels.

The paper is structured as follows: Section \ref{related} describes works related to this task which inspired our
solution, Section \ref{data_descr} describes the data provided by the organizers of the shared task, Section \ref{method} explains our proposed method 
it in detail, Section \ref{evaluation} shows and comments on results of different evaluation metrics that we used for evaluating the performance of our method and of those that the organizers of the shared task used, and in Section \ref{conclusions} we draw our conclusions and open topics for further work.
\section{Related work}
\label{related}

Over the last few years, an increasing number of publications explain different uses of ontologies.

Automatic classification of Web pages based on the concept of domain ontology \cite{1607205} uses an ontology, which expresses terminology information and vocabulary contained in Web documents by way of a hierarchical structure to classify them in real-time and without a learning process. Further, the
Ontology-driven aspect-based sentiment analysis classification~\cite{GARCIADIAZ2020641} also uses an ontology to aid classification by modeling the infectious disease domain with concepts such as risks, symptoms, transmission methods or drugs.

Tax2vec \cite{SKRLJ2020101104} is a data enrichment approach that can use an ontology to form new features from documents, which can be used for learning. 

Using Ontologies and Machine Learning for Hazard Identification and Safety Analysis \cite{usingO} uses ontologies to provide a basis for early identification of system hazards, while Ontologies for Machine Learning \cite{OinML} discusses various uses of ontologies in machine learning.
Similarly to the generalization of financial domain concepts presented in this work, searching for hypernyms of concepts is also discussed in SemEval-2018 Task 9: Hypernym Discovery \cite{ext1} and working with financial domain representations has also been done at the The FinSim 2020 Shared Task \cite{maarouf-etal-2020-finsim}.

\section{Data description}
\label{data_descr}

The dataset consists of one-word or multi-word concepts from the financial domain and their labels, presented in Table \ref{tab:data}. The data is separated into train and test sets.

\begin{table}[H]
    \centering
    
    \begin{tabular}{c|c|c|c}
        Label & Train  \\ \hline
        Forward & 9  \\ \hline
        Funds & 22 \\ \hline
        Future &  19 \\ \hline
        MMIs & 17 \\ \hline
        Option & 24  \\ \hline
        Stocks & 17 \\ \hline
        Swap & 36 \\ \hline
        Equity Index & 286\\ \hline
        Credit Index & 129 \\ \hline
        Bonds & 55  \\ \hline \hline
        Together & 614 \\
    \end{tabular}
    \vspace*{5mm}
    \caption{Distribution of the labels in the train set.}
    \label{tab:data}
\end{table}

\section{Proposed Method}
\label{method}

The proposed method consists of two separate classification approaches. The first tries to find the label associated with the instance concept using The Financial Industry Business Ontology (FIBO), while the other is a Random Forest Classifier \cite{RF} trained on the train set. In the final step, both classifications are merged into one that represents the method's final output.

This section presents classification using an ontology (Section \ref{ontology_based_clf}), classification using a Random Forest Classifier (Section \ref{RF}) and how the outputs of both classifications are merged into the method's final output.

\subsection{Ontology-Augmented Concept Classification}
\label{ontology_based_clf}

We transformed FIBO into a NetworkX MultiGraph \cite{NetworkX} to easily perform graph search. Concepts from the dataset are mapped to the FIBO and then generalized until a valid label (one of those provided by the shared task organizers) is found. Ontology-Augmented Concept Classification does not need a learning process, so the train set was only used to evaluate the performance, before using it on the test set.

All but two labels have a direct representation (a node with the same name) in the ontology. These two are "Equity Index" and "Credit Index". The first has been manually added in the NetworkX graph as the parent of the "Index" node and the second is used as the default label for concepts that cannot be successfully mapped to the ontology or be generalized into a valid label. The decision to manually add "Equity Index" to FIBO and use "Credit Index" as the label for unsuccessfully mapped or generalized concepts has been made by observing the concepts and their respective labels in the train set - a majority of concepts ending with the word "Index" are labeled as "Equity Index", while concepts labeled "Credit Index" do not have a word so frequently represented, and since "Equity Index" was not in the FIBO it made sense to manually add it as the parent of "Index" and use the other as the default label because after this modification most of the unsuccessfully mapped or generalized concepts were labeled as "Credit Index".


\subsubsection{Concept mapping to FIBO}
\label{mapping}

Because the vast majority of concepts in the dataset are not directly represented in the FIBO ontology, a custom mapping had to be made. If the concept has a direct representation in FIBO then its mapping is trivial, otherwise, the concept is split into words it consists of and each word is checked whether it has a representation in the ontology in its singular or plural form. If no such representation is found, there is an identical check for representations for each of the synonyms of these words according to the NLTK Wordnet \cite{wn}. We map and generalize words and their synonyms separately in this order and stop if a word is generalized into a valid label, which means the order in which we choose words from a multi-word concept is important. We assume that the noun of a multi-word concept is the most indicative as to what the concept should be classified as, that is why we choose words in the reverse order - starting with the last word of the concept.

After trying the synonyms in their singular and plural form, if no viable representation is found, the concept is labeled by default as "Credit Index".

\subsubsection{FIBO generalization}
\label{generalization}

Once the concept is mapped to the FIBO ontology, ancestors of the concept are searched iteratively until a valid label is found. An iteration consists of replacing the current set of nodes with their parents and then checking whether any of them is a valid label. When such a label is found, the search stops and the concept is labeled with the found label. If no such label can be found the concept is labeled by default as "Credit Index".

\subsection{Random Forest Classifier}
\label{RF}

Ontology-Augmented Concept Classification described in Section \ref{ontology_based_clf} can only classify concepts, but cannot provide a ranked list of labels requested by the shared task organizers. For that reason, we supplemented  Ontology-Augmented Concept Classification with a Random Forest Classifier.

Concepts from the train and test sets are vectorized by Word2vec \cite{church_2017} provided by the organizers. After the classifier is trained on the vectorized train set it predicts probabilities of labels for each vectorized concept in the test set. Based on these probabilities a ranked list of labels is made.

\subsection{Method's final output}

Each ranked label list acquired by the Random Forest Classifier (from Section \ref{RF}) is modified so that the corresponding predicted label using the Ontology-Augmented Concept Classification \ref{ontology_based_clf} is put in the first place. This modified ranked list of labels is the final output of this method.

\begin{figure}[h]
        \centering
        \resizebox{\columnwidth}{!}{\includegraphics{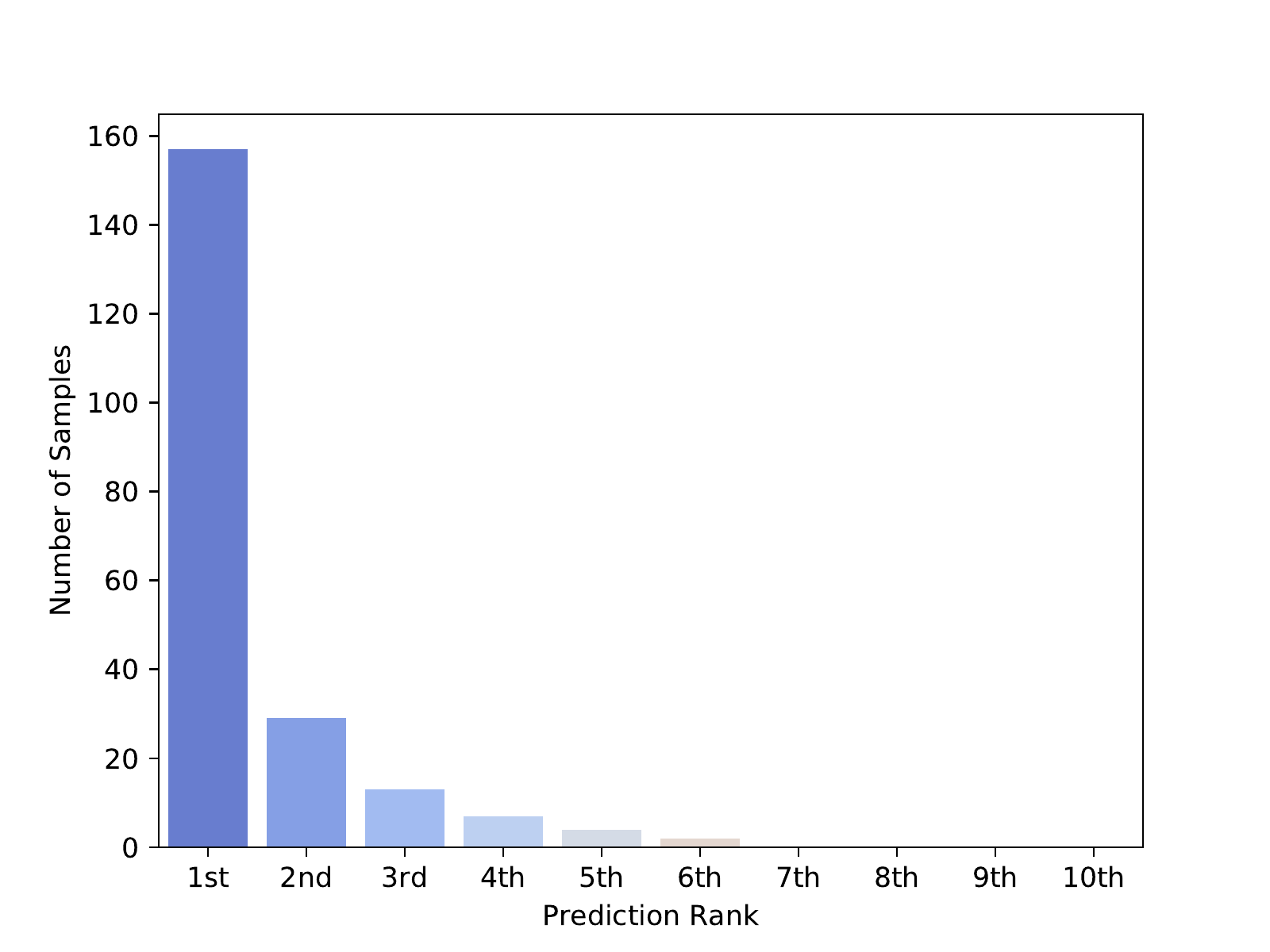}}
        \caption{The graph shows where ontology-based predictions rank on Random Forest's ranked label list.}
        \label{fig:RF_ontology}
\end{figure}

Before modifying the ranked label list and putting the ontology-based prediction in the first place, we analyzed where the latter prediction ranks in the original ranked label list provided by the Random Forrest Classifier. Figure \ref{fig:RF_ontology}. shows how many predictions ranked where on the original ranked label list. 

\section{Evaluation}
\label{evaluation}

Ontology-Augmented Concept Classification explained in Section \ref{ontology_based_clf} was evaluated with accuracy on the train set (Section \ref{accuracy}), since it did not require a learning process, while Random Forest Classifier from Section \ref{RF} did. To evaluate Random Forest Classifier's and the merged method's performance, we have split the train set into two sets: one for training and one for testing (internal test set). Results of the evaluation on this internal test set are shown and discussed in Section \ref{training}.

Official evaluation results of our method on the test set provided by the organizers of the shared task are presented in Section \ref{testing}.



\subsection{Results of the training phase}
\label{training}

This section presents results of the evaluation in the training phase and is structured as follows; the accuracy of Ontology-Augmented Concept Classification is presented in Section \ref{accuracy}, while the accuracy of the Random Forest Classifier is presented in Section \ref{RF_accuracy}. We also present the evaluation of our method using average label rank in Section \ref{average_rank} and in Section \ref{wrongly} we discuss samples from the train set that were not correctly labeled by the Ontology-Augmented Concept Classification.

\subsubsection{Accuracy of Ontology-Augmented Concept Classification}
\label{accuracy}

On the \textbf{train} set Ontology-Augmented Concept Classification has \textbf{0.87} accuracy - accurately predicting 535 out of 614 concepts. 

The shared task organizers provided two baselines for classifying financial concepts. Baseline 1 is a distance-based classifier while Baseline 2 uses logistic regression. The results in Table \ref{tab:accuracy} show that our method outperforms the two baselines provided by organizers of the shared task according to accuracy.

From the two baselines provided by the organizers which output a ranked list of labels we assumed the highest-ranking label as the baseline's prediction and measured the accuracy of both baselines and our Ontology-Augmented Concept Classification.

\begin{table}[H]
    \centering
    \begin{tabular}{c|c}
        Method & Accuracy  \\ \hline
        Baseline 1 (distance) & 0.62  \\ \hline
        Baseline 2 (LR) & 0.86 \\ \hline
        Ontology-Augmented Concept Classification & \textbf{0.87} \\
    \end{tabular}
    \vspace*{5mm}
    \caption{Accuracy of our Ontology-Augmented Concept Classification compared to two baselines provided by the organizers.}
    \label{tab:accuracy}
\end{table}

\subsubsection{Accuracy of Random Forest Classifier}
\label{RF_accuracy}

To evaluate the accuracy of the Random Forest Classifier we have split the train set provided by the organizers into two sets; one containing 90\% of samples and the other 10\%, the latter serving as an internal test set. We have trained the Random Forest Classifier on the larger of the two splits and tested its performance on the smaller one. It predicted with an accuracy of 0.85, while Ontology-Augmented Concept Classification predicted with an accuracy of 0.88 on this same internal test set of data.

\begin{table}[H]
    \centering
    \begin{tabular}{c|c}
        Method & Accuracy  \\ \hline
        Random Forest Classifier & 0.85 \\ \hline
        Ontology-Augmented Concept Classification & 0.88 \\
    \end{tabular}
    \vspace*{5mm}
    \caption{Accuracy of the Random Forest Classifier and the Ontology-Augmented Concept Classification on the internal test set.}
    \label{tab:accuracy_custom_split}
\end{table}

\subsubsection{Average label rank}
\label{average_rank}

Average label rank is a metric that calculates the average rank of the correct label in ranked label lists, which are results of a prediction model. A smaller average rank therefore means a more accurate prediction.

Using the same split of the train set to sets of 90\% and 10\% of data as in Section \ref{RF_accuracy} we evaluated the performance of Random Forest Classifier's output before and after being modified by Ontology-Augmented Concept Classification's predictions, using average label rank.

\begin{table}[H]
    \centering
    \begin{tabular}{c|c}
        Method & Average label rank  \\ \hline
        Random Forest Classifier & 1.37 \\ \hline
        Merged method & 1.19 \\
    \end{tabular}
    \vspace*{5mm}
    \caption{Evaluation of the Random Forest Classifier and the Merged method with an average label rank on the internal test set (10\% of the train set).}
    \label{tab:average_lbl}
\end{table}

While the difference in accuracy between the two methods is not significant as shown in Table \ref{tab:accuracy_custom_split}, results in Table \ref{tab:average_lbl} show that the average label rank is considerably lower after Random Forest's ranked label list is modified by Ontology-Augmented Concept Classification's prediction.

\subsubsection{Wrongly classified concepts}
\label{wrongly}

In this section, we show and explain some examples of financial concepts from the train set that were wrongly classified. These examples can be seen in Table \ref{tab:examples}.

"Agency Bonds" is a concept that does not have a direct representation in FIBO. After failing to find one, Ontology-Augmented Concept Classification then searched for representations of words "Bonds" and "Agency" in this order. "Bonds" is represented in FIBO and is also a valid label, so "Agency Bonds" gets labeled as "Bonds".

"Eurobond" is a one-word concept and is labeled by default as "Credit Index" after failing to find its direct representation in FIBO. Should "Eurobond" be split in two words; "Euro" and "bond", the concept would be classified similarly to the previous example and would be correctly labeled as "Bonds".

As the last example, we look at "Option on Future". In Section \ref{mapping} we assume, that the noun of the multi-word concept is the most indicative word as to what the concept should be labeled as. Because the noun is usually the last word of a multi-word phrase, we choose individual words from the concept in the reverse order (starting with the last word). This example contains two nouns, however. Furthermore, the first noun "Option" is the one leading to the correct classification. Because we choose words in the reverse order we find "Future" as a valid label and classify this concept wrongly. This could be in the future improved by rules determining the head noun of the noun phrase.

\begin{table}[H]
    \centering
    \begin{tabular}{c|c|c}
        Concept  & Predicted label & Label  \\ \hline
        Government Bond Index Linked  & Equity Index & Funds \\ \hline
        Agency Bonds & Bonds & MMIs \\ \hline
        Eurobond  & Credit Index & Bonds \\ \hline
        International depository receipt & Credit Index & Stocks \\ \hline
        Rights & Credit Index & Stocks \\ \hline
        Option on Future & Future & Option \\
    \end{tabular}
    \vspace*{5mm}
    \caption{Examples of concepts that Ontology-Augmented Concept Classification labeled wrongly.}
    \label{tab:examples}
\end{table}

\subsection{Results on the evaluation phase}
\label{testing}

Official results, which were determined on a test set of 212 samples, of the shared task show, that our method had an average label rank of \textbf{1.316}, which ranks 12th out of the 18th approaches, and an accuracy of \textbf{0.811}, which ranks 15th out of the 18th approaches.

\section{Conclusions and further work}
\label{conclusions}

This work presents a combination of a Random Forest Classifier and a synonymy-based ontology matching for concept classification. The method has been developed as a solution for the shared task on Learning Semantic Similarities for the Financial Domain, so the domain ontology is represented by the Financial Industry Business Ontology (FIBO). Our results on the datasets provided by the shared task organizers showcase the ontology's usefulness for the purposes of classification.

We believe further work can be done especially on the mapping of concepts to the domain ontology since this part proved crucial to the successful classification. WordNet synonyms only proved partly successful, we believe it could benefit the method if we could determine domain synonyms of concepts and search for their representations in the ontology. The order in which we choose words from multi-word concepts and determining the position of head nouns can also be further discussed as an approach that ranks words based on their importance for classification might be beneficial.
We developed a solution for the financial domain but this method could be made generic and applicable to any domain provided we have the relevant ontology and the mapping of concepts.

\section{Acknowledgments}
This work was supported by the Slovenian Research Agency (ARRS) grants for the core programme Knowledge technologies (P2-0103) and the project quantitative and qualitative analysis of the unregulated corporate financial reporting (J5-2554 ). The last author was funded through the ARRS junior researcher grant. 

\bibliographystyle{ACM-Reference-Format}
\bibliography{refs}

\end{document}